\documentclass[11pt,letterpaper]{article}
\usepackage{emnlp2017}
\usepackage{times}
\usepackage{latexsym}

\usepackage{hyperref}
\usepackage{url}
\usepackage[utf8]{inputenc} % allow utf-8 input
\usepackage[T1]{fontenc}    % use 8-bit T1 fonts
\usepackage{booktabs}       % professional-quality tables
\usepackage{amsfonts}       % blackboard math symbols
\usepackage{nicefrac}       % compact symbols for 1/2, etc.
\usepackage{microtype}      % microtypography
\usepackage{algorithm,algorithmic}
\usepackage{amsmath}
\usepackage{amssymb}
\usepackage{graphicx}
\usepackage{caption}
\usepackage{subcaption}
\usepackage{amsthm}
\usepackage{bbm}
\usepackage{booktabs}
\usepackage{array,multirow}

\usepackage[utf8]{inputenc}
\usepackage[english]{babel}

\usepackage{newtxmath}
\usepackage{newtxtext}

\emnlpfinalcopy

\def\emnlppaperid{176}

\title{Learning the Dimensionality of Word Embeddings}

\author{Eric Nalisnick \thanks{Authors contributed equally.} \\
Department of Computer Science\\
University of California, Irvine\\
Irvine, CA 92697, USA \\
\texttt{enalisni@uci.edu} \\
\And
Sachin Ravi $^{*}$\\
Department of Computer Science \\
Princeton University \\
Princeton, NJ 08540, USA \\
\texttt{sachinr@cs.princeton.edu} 
}

\date{}

\begin{document}

\maketitle

\begin{abstract}
  We describe a method for learning word embeddings with data-dependent dimensionality.  Our \textit{Stochastic Dimensionality Skip-Gram} (SD-SG) and \textit{Stochastic Dimensionality Continuous Bag-of-Words} (SD-CBOW) are nonparametric analogs of Mikolov et al.'s (2013) well-known `word2vec' models.  Vector dimensionality is made dynamic by employing techniques used by C\^{o}t\'{e} \& Larochelle (2016) to define an RBM with an infinite number of hidden units.  We show qualitatively and quantitatively that the SD-SG and SD-CBOW are competitive with their fixed-dimension counterparts while providing a distribution over embedding dimensionalities, which offers a window into how semantics distribute across dimensions.
\end{abstract}

\section{Introduction}
\textit{Word Embeddings} (WEs) \citep{bengio2003neural, mnih2009scalable, turian2010word, mikolov2013distributed} have received wide-spread attention for their ability to capture surprisingly detailed semantic information without supervision. However, despite their success, WEs still have deficiencies.  One flaw is that the embedding dimensionality must be set by the user and thus requires some form of cross-validation be performed.  Yet, this issue is not just one of implementation.  Words naturally vary in their semantic complexity, and since vector dimensionality is standardized across the vocabulary, it is difficult to allocate an appropriate number of parameters to each word.  For instance, the meaning of the word \textit{race} varies with context (ex: competition vs anthropological classification), but the meaning of \textit{regatta} is rather specific and invariant.  It seems unlikely that \textit{race} and \textit{regatta}'s representations could contain the same number of parameters without one overfitting or underfitting.  

To better capture the semantic variability of words, we propose a novel extension to the popular \textit{Skip-Gram} and \textit{Continuous Bag-of-Words} models \citep{mikolov2013distributed} that allows vectors to have stochastic, data-dependent dimensionality.  By employing the same mathematical tools that allow the definition of an \textit{Infinite Restricted Boltzmann Machine} \citep{cote2015infinite}, we define two log-bilinear energy-based models named \textit{Stochastic Dimensionality Skip-Gram} (SD-SG) and \textit{Stochastic Dimensionality Continuous Bag-of-Words} (SD-CBOW) after their fixed dimensional counterparts.  During training, SD-SG and SD-CBOW allow word representations to grow naturally based on how well they can predict their context.  This behavior, among other things, enables the vectors of specific words to use few dimensions (since their context is reliably predictable) and the vectors of vague or polysemous words to elongate to capture as wide a semantic range as needed.  As far as we are aware, this is the first word embedding method that allows vector dimensionality to be learned and to vary across words. 

\section{Fixed Dimension Word Embeddings}
We first review the original \textit{Skip-Gram} (SG) and \textit{Continuous Bag-of-Words} (CBOW) architectures \citep{mikolov2013distributed} before describing our novel extensions.  In the following model definitions, let $\mathbf{w}_{i} \in \mathbb{R}^{d}$ be a $d$-dimensional, real-valued vector representing the $i$th input word $w_{i}$, and let $\mathbf{c}_{k} \in \mathbb{R}^{d}$ be a vector representing the $k$th context word $c_{k}$ appearing in a $2K$-sized window around an instance of $w_{i}$ in some training corpus $\mathcal{D}$.     

\subsection{Skip-Gram}
SG learns a word's embedding via maximizing the log probability of that word's context (i.e. the words occurring within a fixed-sized window around the input word).  Training the SG model reduces to maximizing the following objective function: \begin{equation}\begin{split}\label{sg_def} \mathcal{L}_{SG} &= \sum_{i=1}^{|\mathcal{D}|} \sum_{i-K\le k \le i+K, k\ne i} \log p(c_{k} | w_{i}) \\ &= \sum_{i=1}^{|\mathcal{D}|} \sum_{i-K\le k \le i+K, k\ne i} \log \frac{e^{\mathbf{c}_{k}^{T}\mathbf{w}_{i}}}{\sum_{v=1}^{V} e^{\mathbf{c}_{v}^{T}\mathbf{w}_{i}}} \end{split}\end{equation} where $V$ is the size of the context vocabulary.  Stochastic gradient descent is used to update not only $\mathbf{w}_{i}$ but also $\mathbf{c}_{k}$ and $\mathbf{c}_{v}$.  A \textit{hierarchical softmax} or \textit{negative sampling} is commonly used to approximate the normalization term in the denominator \citep{mikolov2013distributed}. 
%Negative sampling consists of randomly sampling (either from the uniform or empirical distribution) words to serve as negative examples--words that \textit{do not} appear in the current context window--and then optimizing so that the true context word has a higher likelihood than the samples.  Using negative sampling for computing Equation \ref{sg_def} results in the following modification: $ \log p(c_{k} | w_{i}) \approx \mathbf{c}_{k}^{T}\mathbf{w}_{i} - \log  \left [ \sum_{s=1}^{S} e^{\mathbf{c}_{s}^{T} \mathbf{w}_{i}} + e^{\mathbf{c}_{k}^{T} \mathbf{w}_{i}} \right ]\end{equation} for $S$ negative samples.  Notice that we have defined a slightly different negative sampling objective than proposed in \cite{mikolov2013distributed}, which incorporates logistic regression to discriminate the true context word from the samples.  However, other embedding models such as \cite{huang2013learning} have used the negative samples directly in the normalizing sum, as we do. 

\subsection{Continuous Bag-of-Words}
CBOW can be viewed as the inverse of SG: context words $c_{1}\ldots c_{k}$ serve as input in the prediction of a center word $w_{i}$.  The CBOW objective function is then written as \begin{equation}\begin{split}\label{cbow_def} \mathcal{L}_{CBOW} &= \sum_{i=1}^{|\mathcal{D}|} \log p(w_{i} | c_{i-K}\ldots c_{i+K}) \\ &= \sum_{i=1}^{|\mathcal{D}|} \log \frac{e^{\frac{1}{2K-1}\sum_{j} \mathbf{c}_{j}^{T}\mathbf{w}_{i}}}{\sum_{v=1}^{V} e^{\frac{1}{2K-1}\sum_{j} \mathbf{c}_{j}^{T}\mathbf{w}_{v}}} \end{split}\end{equation} where $\mathbf{c}$, $\mathbf{w}$, and $V$ are defined as above for SG.  Again, the denominator is approximated via negative sampling or a hierarchical softmax.

\section{Word Embeddings with Stochastic Dimensionality}  Having introduced the fixed-dimensional embedding techniques SG and CBOW, we now define extensions that make vector dimensionality a random variable.  In order for embedding growth to be unconstrained, word vectors $\mathbf{w}_{i} \in \mathbb{R}^{\infty}$ and context vectors $\mathbf{c}_{k} \in \mathbb{R}^{\infty}$ are considered infinite dimensional and initialized such that the first few dimensions are non-zero and the rest zero.

\subsection{Stochastic Dimensionality Skip-Gram}
Define the \textit{Stochastic Dimensionality Skip-Gram} (SD-SG) model to be the following joint Gibbs distribution over $\mathbf{w}_{i}$, $\mathbf{c}_{k}$, and a random positive integer $z \in \mathbb{Z}^{+}$ denoting the maximum index over which to compute the embedding inner product: $p(w_{i}, c_{k}, z) = \frac{1}{Z} e^{-E(\mathbf{w}_{i}, \mathbf{c}_{k}, z)}$ where $Z = \sum_{\mathbf{w}} \sum_{\mathbf{c}} \sum_{z} e^{-E(\mathbf{w}, \mathbf{c}, z)}$, also known as the partition function.  Define the energy function as $E(\mathbf{w}_{i}, \mathbf{c}_{k}, z) = z\log a - \sum_{j=1}^{z} w_{i,j}c_{k,j} - \lambda w_{i,j}^{2} - \lambda c_{k,j}^{2} $ where $1 < a < \infty$, $a \in \mathbb{R}$ and $\lambda$ is a weight on the L2 penalty.  Notice that SD-SG is essentially the same as SG except for three modification: (1) the inner product index $z$ is a random variable, (2) an L2 term is introduced to penalize vector length, and (3) an additional $z \log a$ term is introduced to ensure the infinite sum over dimensions results in a convergent geometric series \cite{cote2015infinite}.  This convergence, in turn, yields a finite partition function; see the Appendix for the underlying assumptions and derivation.  

For SD-SG's learning objective, ideally, we would like to marginalize out $z$: \begin{equation}\begin{split}\label{no_l}
\log &p(c_{k} | w_{i}) = \log \sum_{z=1}^{\infty} p(c_{k}, z| w_{i}) \\ &= \log \left [ \sum_{z=1}^{l} p(c_{k}, z| w_{i})  + \frac{a}{a-1}p(c_{k}, l| w_{i})\right]
\end{split}
\end{equation} where $l$ is the maximum index at which $\mathbf{w}$ or $\mathbf{c}$ contains a non-zero value.  $l$ must exist under the sparsity assumption that makes the partition function finite (see Appendix), but that assumption gives no information about $l$'s value.  One work-around is to fix $l$ at some constant, but that would restrict the model and make $l$ a crucial hyperparameter.  

A better option is to sample $z$ values and rely on the randomness to make learning dynamic yet tractable.  This way $l$ can grow arbitrarily (i.e. its the observed maximum sample) while the vectors retain a finite number of non-zero elements.  Thus we write the loss in terms of an expectation \begin{equation}\begin{split}\label{vi_no} &\mathcal{L}_{\text{\tiny{SD-SG}}} = \log p(c_{k} | w_{i}) \\ &= \mathbb{E}_{z|c_{k},w_{i}}\left[ \log p(c_{k}, z| w_{i}) - \log p(z|c_{k},w_{i}) \right].\end{split}\end{equation}  Notice that this is the evidence bound widely used for variational inference except here there is equality, not a bound, because we have set the variational distribution $q(z)$ to the posterior $p(z|w,c)$, which is tractable.  The sampling we desire then comes about via a score function estimate of the gradient: \begin{equation}\label{vi_no2}\begin{split}
&\frac{\partial}{\partial w_{i}} \mathcal{L}_{\text{\tiny{SD-SG}}} \approx \frac{1}{S} \sum_{s=1}^{S} \frac{\partial}{\partial w_{i}} \log p(c_{k}, \hat z_{s}| w_{i}) \\ &+ \left[ \log p(c_{k}| w_{i}) - 1  \right ]\frac{\partial}{\partial w_{i}}\log p(\hat z_{s} | c_{k},w_{i}) \end{split}\end{equation} where $S$ samples are drawn from $\hat z_{s} \sim p(z | c_{k},w_{i})$.  Note the presence of the $p(c_{k}| w_{i})$ term---the very term that we said was problematic in Equation \ref{no_l} since $l$ was not known.  We can compute this term in the Monte Carlo objective by setting $l$ to be the largest $\hat z$ value sampled up to that point in training.  The presence of $p(c_{k}| w_{i})$ is a boon because, since it does not depend $\hat z$, there is no need for control variates to stabilize the typically high variance term $\frac{\partial}{\partial w_{i}}\log p(\hat z_{s} | c_{k},w_{i})$. 

Yet there's still a problem in that $\hat z \in [1, \infty)$ and therefore a very large dimensionality (say, a few thousand or more) could be sampled, resulting in the gradient incurring painful computational costs.  To remedy this situation, if a $\hat z$ value greater than the current value of $l$ is sampled, we set $\hat z = l + 1$, restricting the model to grow only one dimension at a time.  Constraining growth in this way is computationally efficient since $\hat z$ can be drawn from a ($l+1$)-dimensional multinoulli distribution with parameters $\mathbf{\Theta} =  \{\theta_{1}=p(z=1|w,c),\ldots, \theta_{l+1}=\frac{a}{a-1}p(z=l|w,c)  \}$.  The intuition is the model can sample a dimension less than or equal to $l$ if $l$ is already sufficiently large or draw the ($l+1$)th option if not, choosing to increase the model's capacity.  The hyperparameter $a$ controls this growing behavior: as $a$ approaches one from the right, $P(z>l|w)$ approaches infinity.

\subsection{Stochastic Dimensionality Continuous Bag-of-Words}
The \textit{Stochastic Dimensionality Continuous Bag-of-Words} (SD-CBOW) model is a conditional Gibbs distribution over a center word $\mathbf{w}_{i}$ and a random positive integer $z \in \mathbb{Z}^{+}$ denoting the maximum index as before, given multiple context words $\mathbf{c}_{k}$: $p(w_{i}, z | c_{i-K}, \ldots, c_{i+K}) = \frac{1}{Z_{w,z}} e^{-\frac{1}{2K-1}\sum_{j} E(\mathbf{w}_{i}, \mathbf{c}_{j}, z)}$ where $Z_{w,c} = \sum_{\mathbf{w}} \sum_{z} e^{-\frac{1}{2K-1}\sum_{j} E(\mathbf{w}, \mathbf{c}_{j}, z)}$.  The energy function is defined just as for the SD-SG and admits a finite partition function using the same arguments.  The primary difference between SD-SG and SD-CBOW is that the latter assumes all words appearing together in a window have the same vector dimensionality.  The SD-SG, on the other hand, assumes just word-context \emph{pairs} share dimensionality.

Like with SD-SG, learning SD-CBOW's parameters is done via a Monte Carlo objective.  Define the SD-CBOW objective $\mathcal{L}_{\text{\tiny{SD-CBOW}}}$ as \begin{equation}\begin{split}\label{vi_no2_1}
\mathcal{L}_{\text{\tiny{SD-CBOW}}} &= \log p(w_{i} | c_{i-K}\ldots c_{i+K}) \\ &= \mathbb{E}_{z} [\log p(w_{i}, z | c_{i-K}\ldots c_{i+K}) \\ & \qquad -\log p(z|w_{i},c_{i-K}\ldots c_{i+K}) ].
\end{split}
\end{equation}  Again we use a score function estimate of the gradient to produce dynamic vector growth: $
 \frac{\partial}{\partial w_{i}} \mathcal{L}_{\text{\tiny{SD-CBOW}}} \approx \frac{1}{S} \sum_{s=1}^{S} \frac{\partial}{\partial w_{i}} \log p(w_{i}, \hat z_{s}|c_{i-K},\ldots) +\left[ \log p(w_{i} | c_{i-K},\ldots) - 1\right ]\frac{\partial}{\partial w_{i}}\log p(\hat z_{s} | w_{i},c_{i-K},\ldots)$ where $S$ samples are drawn from $\hat z_{s} \sim p(z | w_{i},c_{i-K},\ldots, c_{i+K})$.  Vectors are constrained to grow only one dimension at a time as done for the SD-SG by sampling from a $l+1$th dimensional multinoulli with $\theta_{l+1} = \frac{a}{a-1}p(z = l|w_{i},c_{i-K},\ldots, c_{i+K})$.

\begin{figure*}
\begin{subtable}{.54\textwidth}
\centering
\begin{tabular}{lcc}
\\
& WordSim-353 & MEN \\
\hline \hline
SG-50 & 0.607 & 0.650  \\
CBOW-50 & 0.609 & 0.659  \\
\textbf{SD-CBOW} & 0.614 & 0.660  \\
\textbf{SD-SG} & 0.620 & 0.674  \\ 
CBOW-200 & 0.643 & 0.712  \\
SG-200 & \bf{0.696} & \bf{0.736} \\
\\
\end{tabular}
\caption{Semantic similarity (Spearman's rank correlation)}
\label{table:sim}
\end{subtable}
\begin{subfigure}{.45 \textwidth}
  \centering
  \includegraphics[width=.70\linewidth]{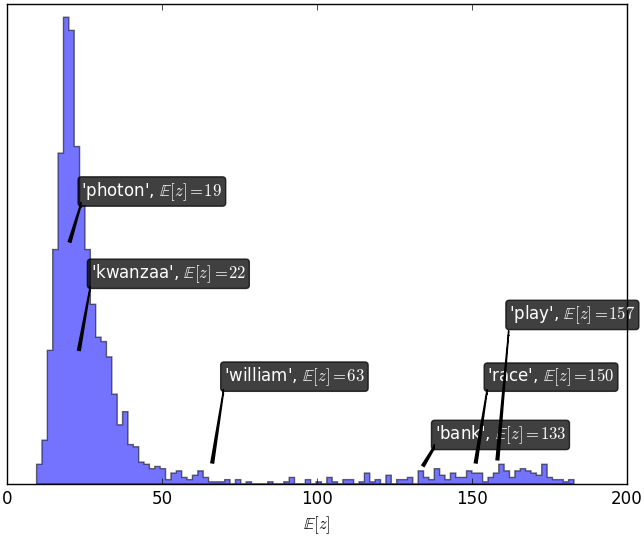}
  \caption{Histogram of the expected dimensionalities.}
\end{subfigure}
\begin{subfigure}{.49\textwidth}
  \centering
  \includegraphics[width=.99\linewidth]{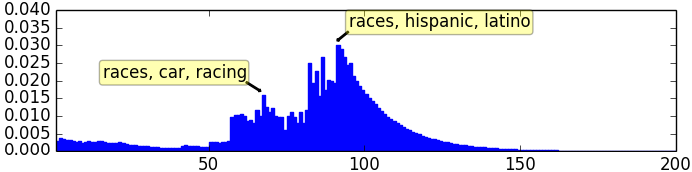}
  \caption{$p(z|w=\text{`race'})$}
  \label{fig:sub1p}
\end{subfigure}
\begin{subfigure}{.49\textwidth}
  \centering
  \includegraphics[width=.99\linewidth]{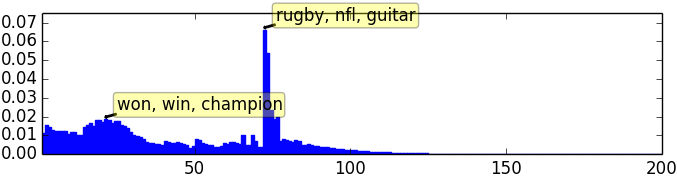}
  \caption{$p(z|w=\text{`player'})$}
  \label{fig:sub2p}
\end{subfigure}
\caption{Subfigure (a) shows results for semantic similarity tasks.  The Spearman's rank correlation between model and human scores are calculated for the WordSim-353 and MEN datasets.  Subfigure (b) shows a histogram of the expected vector dimensionalities after training SD-CBOW.  Subfigures (c) and (d) show the distribution over dimensionalities SD-SG learned for the words \textit{race} and \textit{player}.}
\label{results}
\end{figure*}
\section{Related Work}\label{related}
As mentioned in the Introduction, we are aware of no previous work that defines embedding dimensionality as a random variable whose distribution is learned from data.  Yet, there is much existing work on increasing the flexibility of embeddings via auxiliary parameters.  For instance, Vilnis \& McCallum (2015) represent each word with a multivariate Gaussian distribution.  The Gaussian embedding's mean parameter serves much like a traditional vector representation, and the method's novelty arises from the covariance parameter's ability to capture word specificity \citep{vilnis2014word}.  Other related work proposes using multiple embeddings per word in order to handle \textit{polysemy} and \textit{homonymy} \cite{HuangEtAl2012, reisinger2010multi, neelakantan2015efficient, tian2014probabilistic, bartunov2015breaking}.  Bartunov et al.\ (2016)'s \textit{AdaGram} model is closest in spirit to SD-SG and SD-CBOW in that it uses the Dirichlet Process to learn an unconstrained, data-dependent number of embeddings for each word.  Yet, in contrast to SD-SG/-CBOW, the dimensionality of each embedding is still user-specified.   
%The models mentioned \emph{add} parameters in order to gain capacity and functionality.  SD-SG and SD-CBOW, on the other hand, only contain $2|V| \times z_{\text{max}}$ parameters (where $|V|$ is the vocabulary size and $z_{\text{max}}$ is the max dimension learned), which is the same number as traditional SG and CBOW with a vector size set to $z_{\text{max}}$.  In some sense, SD-SG and SD-CBOW owe their increased functionality to their ability to use \emph{less} parameters, as they can truncate vectors at different dimensions as the current context prescribes.  

\section{Evaluation}\label{eval}
We evaluate SD-SG and SD-CBOW quantitatively and qualitatively against original SG and CBOW.  For all experiments, models were trained on a one billion word subset of Wikipedia (6/29/08 snapshot).  The same learning rate ($\alpha = 0.05$ for CBOW, $\alpha = 0.025$ for SG), number of negative samples (5), context window size (6), and number of training epochs (1) were used for all models.  SD-SG and SD-CBOW were initialized to ten dimensions.

\textbf{Quantitative Evaluation.}  We test each model's ability to rank word pairs according to their semantic similarity, a task commonly used to gauge the quality of WEs. We evaluate our embeddings on two standard test sets: WordSim353 \citep{Finkelstein2001} and MEN \citep{BruniMEN}.  As is typical for evaluation, we measure the Spearman's rank correlation between the similarity scores produced by the model and those produced by humans.  The correlations for all models are reported in Subtable (a) of Figure \ref{results}.  We see that the SD-SG and SD-CBOW perform better than their 50 dimensional counterparts but worse than their 200 dimensional counterparts.  All scores are relatively competitive though, separated by no more than $0.1$. 

\textbf{Qualitative Evaluation.}  Observing that the SD-SG and SD-CBOW models perform comparatively to finite versions somewhere between 50 and 200 dimensions, we qualitatively examine their distributions over vector dimensionalities.  Subfigure (b) of Figure \ref{results} shows a histogram of the expected dimensionality---i.e. $\mathbb{E}_{z|w,c}[z]$---of each vector after training the SD-CBOW model.  As expected, the distribution is long-tailed, and vague words occupy the tail while specific words are found in the head.  As shown by the annotations, the word \textit{photon} has an expected dimensionality of 19 while the homograph \textit{race} has 150.  Note that expected dimensionality correlates with word frequency---due to the fact that multi-sense words, by definition, can be used more frequently---but does not follow it strictly.  For instance, the word \textit{william} is the 482nd most frequently occurring word in the corpus but has an expected length of 62, which is closer to the lengths of much rarer words (around 20-40 dimensions) than to similarly frequent words.

In subfigures (c) and (d) of Figure \ref{results}, we plot the quantity $p(z | w)$ for two homographs, \textit{race} (c) and \textit{player} (d), as learned by SD-SG, in order to examine if their multiple meanings are conspicuous in their distribution over dimensionalities.  For \textit{race}, we see that the distribution does indeed have at least two modes: the first at around 70 dimensions represents car racing, as determined by computing nearest neighbors with that dimension as a cutoff, while the second at around 100 dimensions encodes the anthropological meaning.

\section{Conclusions}
We propose modest modifications to SG and CBOW that allow embedding dimensionality to be learned from data in a probabilistically principled way.  Our models preserve performance on semantic similarity tasks while providing a view--via the distribution $p(z|w,c)$--into how embeddings utilize their parameters and distribute semantic information. 

%\section*{Acknowledgments}
%Do not number the acknowledgment section.
%\clearpage
\bibliography{emnlp2017}
\bibliographystyle{emnlp_natbib}

%\clearpage
\section*{Appendix}
\subsection*{A Finite Partition Function}\label{finite}
\textit{Stochastic Dimensionality Skip-Gram}'s partition function, containing a sum over all countably infinite values of $z$, would seem to be divergent and thus incomputable.  However, it is not, due to two key properties first proposed by \cite{cote2015infinite} to define a \textit{Restricted Boltzmann Machine} with an infinite number of hidden units (iRBM).  They are: \begin{enumerate}
  \item \textbf{Sparsity penalty}: The $L2$ penalty in $E(\mathbf{w}_{i}, \mathbf{c}_{k}, z)$ (i.e. the $w_{i,j}^{2}$ and $c_{k,j}^{2}$ terms) ensures the word and context vectors must have a finite two-norm under iterative gradient-based optimization with a finite initial condition.  In other words, no proper optimization method could converge to the infinite solution if all $\mathbf{w}$ and $\mathbf{c}$ vectors are initialized to have a finite number of non-zero elements \citep{cote2015infinite}.
  \item \textbf{Per-dimension constant penalty}:  The energy function's $z\log a$ term results in dimensions greater than $l$ becoming a convergent geometric series.  This is discussed further below.  
\end{enumerate}
With those two properties in mind, consider the conditional distribution of $z$ given an input and context word: \begin{equation}
p(z | w, c) = \frac{e^{-E(\mathbf{w}, \mathbf{c}, z)}}{\sum_{z'=1}^{\infty} e^{-E(\mathbf{w}, \mathbf{c}, z')}}. \end{equation}  Again, the denominator looks problematic due to the infinite sum, but notice the following: \begin{equation}\begin{split}\label{z_finite}
Z_{z} &= \sum_{z'=1}^{l} e^{-E(\mathbf{w}, \mathbf{c}, z')} + \sum_{z'=l+1}^{\infty} e^{-E(\mathbf{w}, \mathbf{c}, z')} \\ &= \sum_{z'=1}^{l} e^{-E(\mathbf{w}, \mathbf{c}, z')} + e^{-E(\mathbf{w}, \mathbf{c}, l)} \sum_{z'=0}^{\infty} \frac{1}{a^{  z' }} \\ &= \sum_{z'=1}^{l} e^{-E(\mathbf{w}, \mathbf{c}, z')} + \frac{a}{a-1}e^{-E(\mathbf{w}, \mathbf{c}, l)}.
\end{split}
\end{equation}  The sparsity penalty allows the sum to be split as it is in step \#2 into a finite term ($\sum_{z'=1}^{l} e^{-E(\mathbf{w}, \mathbf{c}, z')}$) and an infinite sum ($\sum_{z'=l+1}^{\infty} e^{-E(\mathbf{w}, \mathbf{c}, z')}$) at an index $l$ such that $w_{i,j}=c_{k,j}=0$ $\ \forall j>l$.  After $e^{-E(\mathbf{w}, \mathbf{c}, l)}$ is factored out of the second term, all remaining $w_{i,j}$ and $c_{k,j}$ terms are zero.  A few steps of algebra then reveal the presence of a convergent geometric series.  Intuitively, we can think of the second term, $\frac{a}{a-1}e^{-E(\mathbf{w}, \mathbf{c}, l)}$, as quantifying the data's need to expand the model's capacity given $w$ and $c$.   

\end{document}